\documentclass[a4paper,fleqn]{cas-sc}

\usepackage[numbers, sort&compress]{natbib}
\usepackage{caption}

\begin{document}
\let\WriteBookmarks\relax
\def\floatpagepagefraction{1}
\def\textpagefraction{.001}

\newcommand{\paperTitle}{Machine learning approach to brain tumor detection and classification}

%\shorttitle{Machine Learning Approach to Brain Tumor Detection and Classification}
\shorttitle{\paperTitle}
%\shortauthors{First Author et~al.}
%\shortauthors{Alice Oh, Inyoung Noh, Jian Choo, Jihoo Lee, Justin Kim, Justin Park, Kate Hwan}
\shortauthors{AO, IN, JC, JL, JP, KH, SK, and SMO}
%\begin{frontmatter}

%\title [mode = title]{Machine Learning Approach to Brain Tumor Detection and Classification}
\title [mode = title]{\paperTitle}
\tnotemark[1]

\tnotetext[1]{This research was conducted under the supervision of Dr. Soomin Oh (Author name: Soo Min Oh (S.M.O.), e-mail:\href{mailto:soominoh@mit.edu}{soominoh@mit.edu}), who is the corresponding author. All other authors (A.O., I.N., J.C., J.L., J.P., K.H., and S.K.) are listed in alphabetical order and contributed equally to this work.}

\author[1]{Alice Oh}[
%                        type=editor,
%                        auid=000,bioid=1,
%                        prefix=Sir,
%                        role=Researcher,
%                        orcid=0000-0001-0000-0000]
                        ]
%\cormark[1] % considered as a corresponding author
%\fnmark[1] % add a footnote
%\ead{daeun.oh9318@g.gcpsk12.org}
%\ead[url]{www.jkkrishnan.in}
\credit{Investigation, Software, Writing - review \& editing}
\affiliation[1]{organization={Peachtree Ridge High School},
%                addressline={}, 
                city={Duluth},
                citysep={}, % Uncomment if no comma needed between city and postcode
                postcode={30097}, 
                state={GA},
                country={USA}}

\author[2]{Inyoung Noh}[]
%\ead{gracesh2636.gmail.com}
\credit{Investigation, Software, Writing - review \& editing}
\affiliation[2]{organization={Everlove Christian School},
                city={Daejeon},
                citysep={}, % Uncomment if no comma needed between city and postcode
                postcode={34084}, 
%                state={},
                country={Korea}}
                
\author[3]{Jian Choo}[]
%\ead{jian.choo4709@g.gcpsk12.org}
\credit{Investigation, Software, Writing - review \& editing}
\affiliation[3]{organization={North Gwinnett High School},
                city={Suwanee},
                citysep={}, % Uncomment if no comma needed between city and postcode
                postcode={30024}, 
                state={GA},
                country={USA}}

\author[4]{Jihoo Lee}
%\ead{jihoo.hanspharm@gmail.com}
\credit{Investigation, Software, Writing - review \& editing}
\affiliation[4]{organization={Horizon Christian Academy},
                city={Dawsonville},
                citysep={}, % Uncomment if no comma needed between city and postcode
                postcode={30534}, 
                state={GA},
                country={USA}}
                
\author[5]{Justin Park}
%\ead{justin.park7252@g.gcpsk12.org}
\credit{Investigation, Software, Writing - review \& editing}
\affiliation[5]{organization={Seckinger High School},
                city={Buford},
                citysep={}, % Uncomment if no comma needed between city and postcode
                postcode={30519}, 
                state={GA},
                country={USA}}

%\author[6]{Kate Hwang}
\author[3]{Kate Hwang}
%\ead{kate.hwang2095@g.gcpsk12.org}
\credit{Investigation, Software, Writing - review \& editing}
%\affiliation[6]{organization={North Gwinnett High School},
%                city={Suwanee},
%                citysep={}, % Uncomment if no comma needed between city and postcode
%                postcode={30024}, 
%                state={GA},
%                country={USA}}

\author[6]{Sanghyeon Kim}
%\ead{justinkim2510@gmail.com}
\credit{Investigation, Software, Writing - review \& editing}
\affiliation[6]{organization={Horizon Christian Academy },
                city={Cumming},
                citysep={}, % Uncomment if no comma needed between city and postcode
                postcode={30028}, 
                state={GA},
                country={USA}}

\author[7,8]{Soo Min Oh}[
%                        type=editor,
%                        auid=000,bioid=1,
%                        prefix=Sir,
%                        role=Researcher,
%                        orcid=0000-0001-0000-0000]
%                        orcid=0000-0003-2186-1232]
                        ]
\cormark[1] % considered as a corresponding author
\cortext[cor1]{Corresponding author}
%\fnmark[1] % add a footnote
\ead{soominoh@mit.edu}
%\ead[url]{www.jkkrishnan.in}
\credit{Conceptualization, Formal analysis, Funding acquisition, Investigation, Methodology, Software, Supervision, Writing - original draft, Writing - review \& editing}
\affiliation[7]{organization={Wireless Information and Network Sciences Laboratory, Massachusetts Institute of Technology},
%                addressline={}, 
                city={Cambridge},
                citysep={}, % Uncomment if no comma needed between city and postcode
                postcode={02139}, 
                state={MA},
                country={USA}}
\affiliation[8]{organization={Laboratory for Information and Decision Systems, Massachusetts Institute of Technology},
%                addressline={}, 
                city={Cambridge},
                citysep={}, % Uncomment if no comma needed between city and postcode
                postcode={02139}, 
                state={MA},
                country={USA}}

%\cortext[cor1]{Corresponding author}
%\cortext[cor2]{Principal corresponding author}
%\fntext[fn1]{This is the first author footnote, but is common to third author as well.}
%\fntext[fn2]{Another author footnote, this is a very long footnote and it should be a really long footnote. But this footnote is not yet sufficiently long enough to make two lines of footnote text.}

%\nonumnote{This note has no numbers. In this work we demonstrate $a_b$ the formation Y\_1 of a new type of polariton on the interface between a cuprous oxide slab and a polystyrene micro-sphere placed on the slab.}

\begin{abstract}
Brain tumor detection and classification are critical tasks in medical image analysis, particularly in early-stage diagnosis, where accurate and timely detection can significantly improve treatment outcomes. In this study, we apply various statistical and machine learning models to detect and classify brain tumors using brain MRI images. We explore a variety of statistical models including linear, logistic, and Bayesian regressions, and the machine learning models including decision tree, random forest, single-layer perceptron, multi-layer perceptron, convolutional neural network (CNN), recurrent neural network, and long short-term memory. Our findings show that CNN outperforms other models, achieving the best performance. Additionally, we confirm that the CNN model can also work for multi-class classification, distinguishing among four categories of brain MRI images such as normal, glioma, meningioma, and pituitary tumor images. This study demonstrates that machine learning approaches are suitable for brain tumor detection and classification, facilitating real-world medical applications in assisting radiologists with early and accurate diagnosis.
\end{abstract}

%\begin{graphicalabstract}
%\includegraphics{figs/cas-grabs.pdf}
%\end{graphicalabstract}

%\begin{highlights}
%\item Research highlights item 1
%\item Research highlights item 2
%\item Research highlights item 3
%\end{highlights}

\begin{keywords}
Machine learning \sep Brain tumor \sep Neural networks \sep Detection \sep Multi-class classification
\end{keywords}

\maketitle

\section{Introduction}

%\begin{figure}
%	\centering
%	\includegraphics[width=.9\textwidth]{figs/cas-munnar-2024.jpg}
%	\caption{The beauty of Munnar, Kerala. (See also Table \protect\ref{tbl1}).}
%	\label{FIG:1}
%\end{figure}

Brain tumors are among the most life-threatening conditions, and early detection is critical for improving patient outcomes. A brain tumor arises from abnormal growth of cells in the brain, which can affect vital cognitive and motor functions. There are more than $100$ different types of brain tumors~\cite{ostrom_epidemiology_2021,komori_2021_2022,delaidelli_recent_2024}, including benign (non-cancerous) and malignant (cancerous) tumors. According to the medical observation, gliomas, meningiomas, and pituitary tumors are the most common and account for approximately $45\%$, $15\%$, and $15\%$ of all brain tumors, respectively~\cite{yildirim_detection_2023,haq_dacbt_2022}. Malignant tumors are aggressive and can invade surrounding brain tissues, leading to neurological damage or death if not detected early. 
Traditional diagnostic methods for brain tumors rely heavily on manual interpretation of magnetic resonance imaging (MRI) scans by radiologists~\cite{veiga-canuto_comparative_2022,kaur_extracting_2023,magboo_classification_2025}. MRI is considered the gold standard for brain tumor detection due to its high resolution and ability to visualize soft tissues in detail . However, manual diagnosis of brain tumors from MRI is not only time-consuming but also prone to human error due to factors such as radiologist fatigue and variability in expertise. Studies have shown that inter-observer variability in tumor diagnosis can lead to inconsistencies in classification, particularly for less common tumors.

In recent years, advances in machine learning have transformed the field of medical diagnostics. Machine learning algorithms, particularly deep learning models such as Convolutional Neural Networks (CNNs)~\cite{yildirim_detection_2023,srinivasan_hybrid_2024}, have shown remarkable potential in analyzing medical images for various tasks, including the classification of brain tumors. These models can automatically learn features from raw MRI data, reducing the need for manual feature extraction, which is a key limitation of traditional machine learning methods. CNNs, for instance, have been shown to outperform traditional methods in multiple studies, achieving high accuracy rates in detecting and classifying brain tumors. Additionally, techniques such as transfer learning~\cite{ullah_effective_2022} have further enhanced the accuracy of CNN models by allowing pre-trained models to be fine-tuned on medical datasets, thus reducing the need for large amounts of labeled data.

In this study, we compare various statistical and machine learning models~\cite{sievering_comparison_2022,shah_comparative_2020,couronne_random_2018,zeghdaoui_medical-based_2021,petmezas_automated_2022,shiri_comprehensive_2023, oh_machine_2023,song_exploring_2023}, including linear regression (LR), logistic regression (LogR), Bayesian regression (BR), decision tree (DT), random forest (RF), support vector machine (SVM), single-layer perceptron (SLP), multi-layer perceptron (MLP), CNN, recurrent neural network (RNN), and long short-term memory (LSTM), to identify the best model for brain tumor detection and classification. We aim to demonstrate how machine learning, particularly CNNs, can outperform traditional methods and assist in real-world medical applications. In addition to binary classification, we extend our analysis to multi-class classification, distinguishing among normal brain MRI images and three types of brain tumor MRI images including glioma, meningioma, and pituitary tumor images.

This paper is organized as follows: In section~\ref{sec:Data Description}, we introduce brain MRI images considered in this work. Subsequently, we describe the methods including dataset preparation, machine learning approaches, and evaluation metrics in section~\ref{sec:Methods}. The corresponding outcomes are presented in section~\ref{sec:Results}. Finally, we mention the limitations of this work in section~\ref{sec:Limitations} and conclude our work in section~\ref{sec:Conclusions}.

\section{Data Description}
\label{sec:Data Description}
\begin{figure*}[!t]
\centering
\includegraphics[width=.24\textwidth]{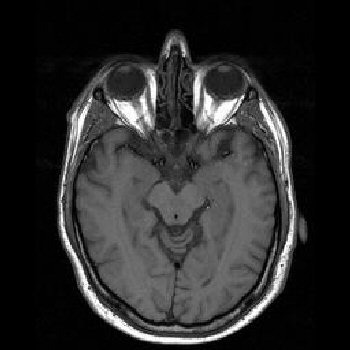}%
    \hspace{0.02cm}
    \includegraphics[width=.24\textwidth]{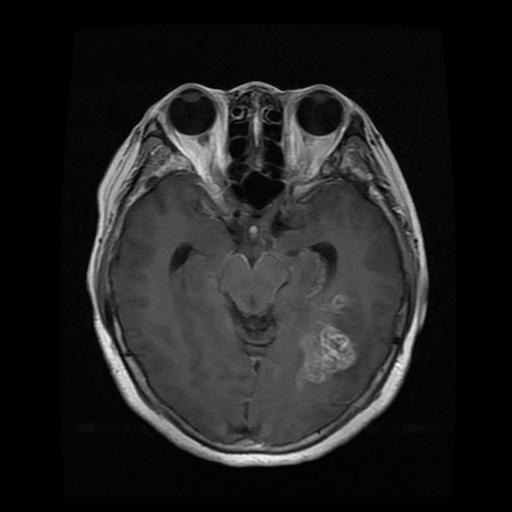}%
    \hspace{0.02cm}
    \includegraphics[width=.24\textwidth]{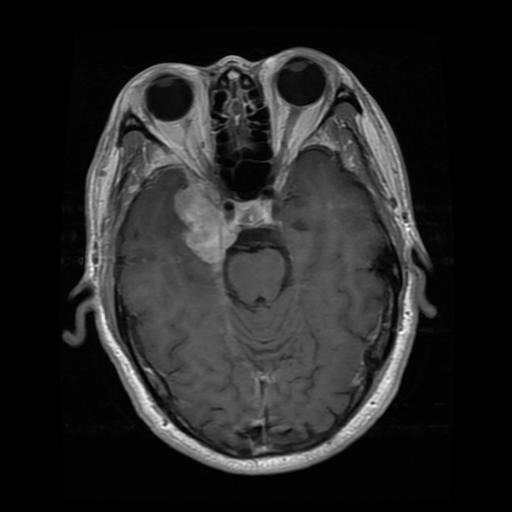}%
    \hspace{0.02cm}
    \includegraphics[width=.24\textwidth]{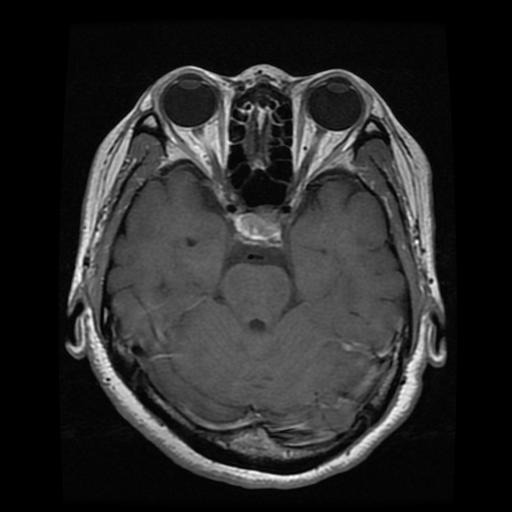}
\caption{The brain MRI images of normal tissues, gliomas, meningiomas, and pituitary tumors from left to right}
\label{FIG:Brain_Tumore_Images}
\end{figure*}
We utilized a publicly available dataset from Kaggle~\cite{sartaj_bhuvaji_ankita_kadam_prajakta_bhumkar_sameer_dedge_swati_kanchan_2020}, which includes MRI images for brain tumor classification. The dataset used in this study consists of $3264$ brain MRI images including $2870$ images for training dataset and $394$ images for test dataset. Each image is labeled with one of four categories such as normal, glioma, meningioma, and pituitary tumor images as shown in Figure~\ref{FIG:Brain_Tumore_Images}. The training dataset includes $395$ normal, $826$ glioma, $822$ meningioma, and $827$ pituitary tumor images. The test dataset contains $105$ normal, $100$ glioma, $115$ meningioma, and $74$ pituitary tumor images. MRI is widely recognized as one of the most effective imaging modalities for detecting soft tissue anomalies, including brain tumors.

\section{Methods}
\label{sec:Methods}

\subsection{Dataset Preparation}

For preprocessing, the grayscale values were extracted from each image, and all images were resized to a uniform resolution of $32\times32$ pixels to ensure consistency in the input data for the model. The pixel intensity values of each image were normalized to the range $[0, 1]$. The training dataset was divided into two subsets such as training ($80\%$) and validation ($20\%$).
For the detection which is the binary classification task, the normal images were labeled as $0$ and all the tumor images were labeled as $1$. After applying one-hot encoding, $0$ and $1$ were encoded to $(1,0)$ and $(0,1)$, respectively. For the multi-class classification task, the normal, glioma, meningioma, and pituitary tumor images were labeled as $0$, $1$, $2$, and $3$, respectively. After one-hot encoding process, $0$, $1$, $2$, and $3$ are encoded to $(1,0,0,0)$, $(0,1,0,0)$, $(0,0,1,0)$, and $(0,0,0,1)$, respectively. Each image will be classified into one of such four categories.

\subsection{Machine Learning Approaches}
We apply three statistical models and eight machine learning models to the brain tumor detection task to classify normal and tumor images. The statistical models include LR, LogR, and BR, and the machine learning models include DT, RF, SVM, SLP, MLP, CNN, RNN, and LSTM.
LR is a statistical method used to model the relationship between a dependent variable and one or more independent variables. LogR is used for binary or multi-class classification tasks by modeling the probability of categorical outcomes. BR incorporates prior information to provide probabilistic predictions.
DT is a simple model that splits data into subsets based on the most informative features. RF is an ensemble of decision trees that enhances prediction accuracy and reduces overfitting. SVM is a supervised learning algorithm that finds the optimal decision boundary in a high-dimensional space for classifying data. SLP is a basic neural network model used for linear classification. MLP is a deeper neural network capable of capturing more complex relationships. CNN is highly effective for image data, leveraging convolutional layers to capture spatial features. RNN is designed for sequential data, where the current input depends on previous inputs. LSTM is a special type of RNN that can remember long-term dependencies in data.
The CNN is expected to perform well on the MRI dataset due to its ability to capture spatial features. Therefore, we additionally apply the CNN to the multi-class classification task to classify four categories such as normal, glioma, meningioma, and pituitary tumor images.

In the SLP, MLP, CNN, RNN, and LSTM, the activation functions of the output layer and the other layers are the softmax~\cite{goodfellow_deep_2016} and scaled exponential linear unit (SELU)~\cite{klambauer_self-normalizing_2017}, respectively. The loss function is the binary cross-entropy for the detection task and the categorical cross-entropy for the multi-class classification task. The Adam optimizer~\cite{kingma_adam_2014} is used to train the machines. The programming was implemented mainly using TensorFlow~\cite{abadi_tensorflow_2016} and Scikit-learn libraries~\cite{pedregosa_scikit-learn_2011}.

\subsection{Evaluation Metrics for Model Performance}
The performance of the models is evaluated using several key metrics such as accuracy, precision, recall, and F1 score. These metrics are widely used in machine learning to measure the performance of classification models, particularly in medical image analysis. Accuracy measures the proportion of correctly classified instances out of the total number of instances. It is calculated using the following formula
\begin{equation}
\text{Accuracy} = \frac{TP + TN}{TP + TN + FP + FN},
\end{equation}
where $TP$, $TN$, $FP$, and $FN$ denote true positives for correctly classified tumor cases, true negatives for correctly classified normal cases, false positives for normal cases incorrectly classified as tumor, false negatives for tumor cases incorrectly classified as normal, respectively. Precision measures the ratio of correctly predicted positive observations to the total predicted positives. It is a measure of the accuracy of the positive predictions made by the model, and written as
\begin{equation}
\text{Precision} = \frac{TP}{TP + FP}.
\end{equation}
Recall (also known as sensitivity or true positive rate) measures the ratio of correctly predicted positive observations to all observations in the actual class (true positives and false negatives)
\begin{equation}
\text{Recall} = \frac{TP}{TP + FN}.
\end{equation}
The F1 score is the harmonic mean of precision and recall and written as
\begin{equation}
%\text{F1 Score} = 2 \times \frac{\text{Precision} \times \text{Recall}}{\text{Precision} + \text{Recall}}.
\text{F1 score} = \frac{2}{\text{Precision}^{-1} + \text{Recall}^{-1}}.
\end{equation}
It is useful when there is an uneven class distribution or when both precision and recall are important.

%\begin{table}[width=0.45\linewidth,cols=5,pos=h]
%\begin{table}[width=0.55\linewidth,cols=5,pos=h]
\begin{table}[width=1.0\linewidth,cols=5,pos=h]
\caption{Accuracy, precision, recall and F1 scores for linear regression (LR), logistic regression (LogR), Bayesian regression (BR), decision tree (DT), random forest (RF), support vector machine (SVM), single-layer perceptron (SLP), multi-layer perceptron (MLP), convolutional neural network (CNN), recurrent neural network (RNN), and  long short-term memory (LSTM) on the test dataset for the brain tumor detection task. The best performance with the CNN is highlighted in bold.}
\begin{tabular*}{\tblwidth}{@{} LLLLL@{} }
\hline
    & \multicolumn{4}{c}{Test}                             \\
    ~~~~~~~~~~~& Accuracy & Precision & Recall  & F1 score \\ \hline
LR  & 80.71\%  & 96.51\%   & 76.47\% & 85.33\%  \\
LogR& 85.28\%  & 97.53\%   & 82.01\% & 89.10\%  \\
BR& 80.20\%  & 93.06\%   & 78.89\% & 85.39\%  \\
DT & 74.37\%  & 97.00\%   & 67.13\% & 79.35\%  \\
RF & 75.38\%  & 97.06\%   & 68.51\% & 80.32\%  \\
SVM & 85.53\%  & 94.27\%   & 85.47\% & 89.66\%  \\
SLP & 85.28\%  & 96.76\%   & 82.70\% & 89.18\%  \\
MLP & 86.04\%  & 96.80\%   & 83.74\% & 89.80\%  \\
\textbf{CNN} & \textbf{91.88\%}  & \textbf{99.23\%}   & \textbf{89.62\%} & \textbf{94.18\%}  \\
RNN & 82.99\%  & 95.49\%   & 80.62\% & 87.43\%  \\ 
LSTM & 86.80\%  & 97.98\%   & 83.74\% & 90.30\%  \\ \hline
\end{tabular*}
\label{tab:Detection_Results}
\end{table}

\begin{figure}[]
\centering
\includegraphics[width=.5\textwidth]{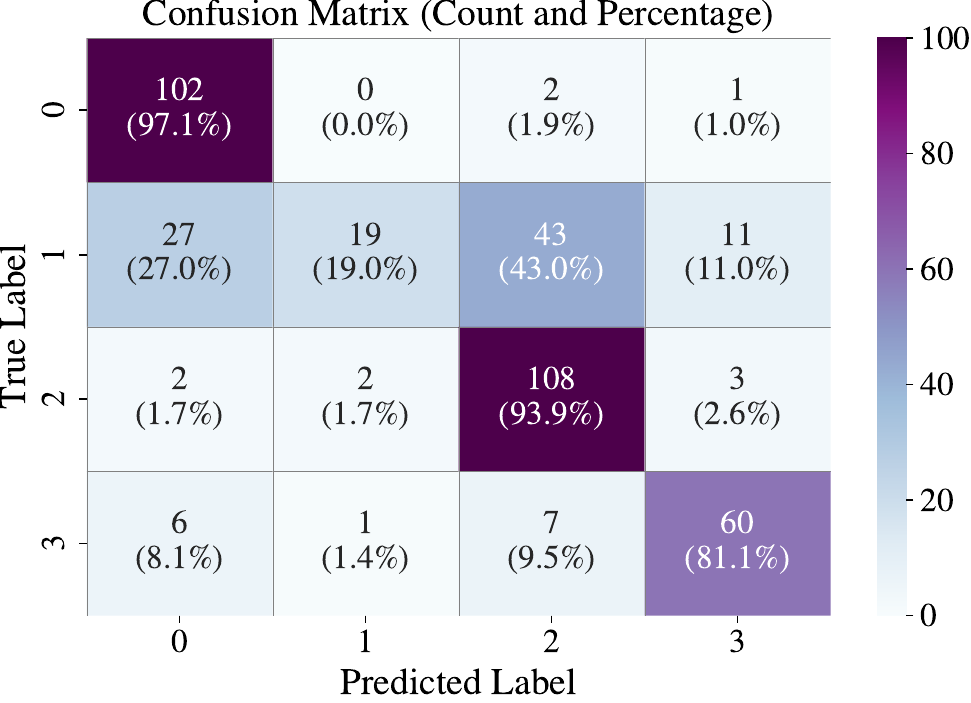}
\caption{Confusion matrix for the trained CNN in classifying normal, glioma, meningioma, and pituitary tumor images, labeled as 0, 1, 2, and 3, respectively, on the test dataset for multi-class classification of brain MRI images.
}
\label{FIG:Confusion_Matrix_for_Multi-Class_Classification}
\end{figure}

\section{Results}
\label{sec:Results}

We first trained the LR, LogR, BR, SVM, SLP, MLP, CNN, RNN, and LSTM for the brain tumor detection which is the binary classification task. Table~\ref{tab:Detection_Results} presents the corresponding accuracy, precision, recall and F1 scores on the test dataset. Among the evaluated models, the CNN achieved the best overall performance, with the accuracy of $91.88\%$, the precision of $99.23\%$, the recall of $89.62\%$, and the F1 score of $94.18\%$. It confirms that CNN can capture spatial features well in MRI images compared to other models. 

Based on these results, we also trained the CNN for the multi-class classification to classify normal, glioma, meningioma, and pituitary tumor images. The confusion matrix shown in Figure~\ref{FIG:Confusion_Matrix_for_Multi-Class_Classification} shows a detailed visualization of the CNN model's outcomes for multi-class classification tasks involving normal images and three types of brain tumors on the test dataset. Table~\ref{tab:Multi-Class_Classification_Results} also presents all evaluation metrics per class for the multi-class classification tasks on the test dataset. 
All outcomes confirm that the model performs well in classifying normal brain tissues with an accuracy of $97.14\%$, meningiomas with $93.91\%$, and pituitary tumors with $81.08\%$, demonstrating its strong capability in detecting and classifying these types. However, classifying gliomas remains challenging, with an accuracy of just $19.00\%$. The low performance for gliomas suggests that many instances of them are not detected correctly, which is likely due to their infiltrative nature and the way they blend with surrounding brain tissue~\cite{grogan_clinical_2022,al-adli_advances_2023,nicholson_diffuse_2021,verburg_state---art_2021,verburg_improved_2020}, emphasizing that gliomas are more diffuse and challenging to classify accurately using traditional imaging techniques. This indicates that despite strong performance for most categories, further improvements are needed for detecting and classifying gliomas.
%
%These challenges are consistent with the literature, which highlights gliomas as being more diffuse and difficult to segment accurately using traditional imaging techniques.
%
%\begin{table}[width=0.55\linewidth,cols=5,pos=h]
\begin{table}[width=1.0\linewidth,cols=5,pos=h]
\caption{Accuracy, precision, recall, and F1 scores per class for the trained CNN in classifying normal, glioma, meningioma, and pituitary tumor images on the test dataset for multi-class classification of brain MRI images.}
\begin{tabular*}{\tblwidth}{@{} LLLLL@{} }
\hline
    & \multicolumn{4}{c}{Test}                             \\
    ~~~~~~~~~~~& Accuracy & Precision & Recall  & F1 score \\ \hline
%CNN_7_for_Multi_Class
%Overall      & 70.56\%  & 76.55\%   & 69.76\% & 65.92\%  \\
%Normal      & 97.14\%  & 62.58\%   & 97.14\% & 76.12\%  \\
%Glioma      & 18.00\%  & 90.00\%   & 18.00\% & 30.00\%  \\
%Meningioma  & 89.57\%  & 71.53\%   & 89.57\% & 79.54\%  \\
%Pituitary tumor  & 74.32\%  & 82.09\%   & 74.32\% & 78.01\%  \\ \hline
%CNN_8_for_Multi_Class
%Overall     & 71.32\%  & 74.21\%   & 70.91\% & 67.04\%  \\
%Normal      & 93.33\%  & 68.06\%   & 93.33\% & 78.72\%  \\
%Glioma      & 21.00\%  & 84.00\%   & 21.00\% & 33.60\%  \\
%Meningioma  & 89.57\%  & 71.03\%   & 89.57\% & 79.23\%  \\
%Pituitary tumor  & 79.73\%  & 73.75\%   & 79.73\% & 76.62\%  \\ \hline
%CNN_9_for_Multi_Class
%Overall     & 73.35\%  & 77.08\%   & 72.78\% & 68.63\%  \\
Normal      & 97.14\%  & 74.45\%   & 97.14\% & 84.30\%  \\
Glioma      & 19.00\%  & 86.34\%   & 19.00\% & 31.15\%  \\
Meningioma  & 93.91\%  & 67.50\%   & 93.91\% & 78.55\%  \\
Pituitary tumor   & 81.08\%  & 80.00\%   & 81.08\% & 80.54\%  \\ \hline
%Wavelet_Type6_CNN_Ave_2_for_Multi-Class
%Overall     & 73.10\%  & 77.80\%   & 72.72\% & 68.89\%  \\
%Normal      & 93.33\%  & 68.53\%   & 93.33\% & 79.03\%  \\
%Glioma      & 19.00\%  & 86.34\%   & 19.00\% & 31.15\%  \\
%Meningioma  & 94.78\%  & 68.99\%   & 94.78\% & 79.85\%  \\
%Pituitary tumor   & 83.78\%  & 87.32\%   & 83.78\% & 85.52\%  \\ \hline
\end{tabular*}
\label{tab:Multi-Class_Classification_Results}
\end{table}

\section{Limitations}
\label{sec:Limitations}
Our study presents several limitations. The research largely focused on reviewing existing statistical and machine learning methods for brain tumor detection and classification rather than contributing entirely new approaches. The medical images used in the study were sourced from publicly available datasets on Kaggle, which may not fully represent the diversity and complexity found in real-world medical cases. We resized the brain MRI images to a low resolution of $32\times32$ pixels to reduce computational costs, which may have compromised the accuracy of the models due to a loss of image details. Additionally, the low performance in classifying gliomas highlights the difficulty in identifying these tumors accurately due to their infiltrative nature and how they blend with surrounding tissue.
To overcome these limitations, we plan to use more exclusive and complex datasets, such as those obtained from medical institutions, to improve the novelty and real-world applicability of the study. Higher-resolution images will be used to preserve important image details, improving the accuracy of the models. Additionally, we will explore advanced machine learning techniques, such as wavelet-based neural networks~\cite{khatami_convolutional_2020,bagaria_wavelet_2021,de_santana_deep-wavelet_2022}, transfer learning~\cite{ullah_effective_2022,matsoukas_what_2022,kim_transfer_2022}, transformers~\cite{shamshad_transformers_2023,usman_analyzing_2022}, large language models~\cite{lee_biobert_2020,huang_clinicalbert_2020}, and quantum machine learning~\cite{wei_quantum_2023,flother_state_2023,peral-garcia_systematic_2024,solenov_potential_2018,ur_rasool_quantum_2023,maheshwari_quantum_2022,zeguendry_quantum_2023,ullah_quantum_2024}, which have shown significant promise in the field of medical imaging.

\section{Conclusions}
\label{sec:Conclusions}
This study compared various statistical and machine learning models for brain tumor detection and classification on brain MRI images. Among the evaluated models, CNN outperformed other models in brain tumor detection, achieving high accuracy, precision, recall, and F1 scores. While other statistical and machine learning models provided moderate performance, the deep learning capability of CNN allowed for superior spatial feature extraction and learning from complex brain MRI images. Additionally, this study also highlights the applicability of CNN for multi-class classification in distinguishing among normal brain MRI images and three types of brain tumor MRI images including glioma, meningioma, and pituitary tumor images.
Machine learning approaches not only show promise in assisting radiologists with early and accurate diagnosis but also provide opportunities for optimizing diagnostic processes through automation. Future work may involve refining CNN architectures, incorporating additional imaging modalities, and validating the models on larger datasets for improved generalization across different clinical settings.

\section*{Acknowledgements}
This research was conducted under the supervision of S.M.O, who is the corresponding author. All other authors (A.O., I.N., J.C., J.L., J.P., K.H., and S.K.) contributed equally to this work and are listed in alphabetical order. This work was supported by the National Research Foundation of Korea with Grant No. NRF-2021R1A6A3A01087148 (S.M.O.).

\section*{Data Availability Statement}

The data and source code that support the findings of this study are available from the corresponding author upon reasonable request.

%\appendix
%\section{My Appendix}
%Appendix sections are coded under \verb+\appendix+.

\printcredits

%% Loading bibliography style file
%\bibliographystyle{model1-num-names}
%\bibliographystyle{cas-model2-names}
%\bibliographystyle{plain}
%\bibliographystyle{unsrt}
%\bibliographystyle{elsarticle-num}

% Loading bibliography database
%\bibliography{cas-refs}
%\bibliography{references.bib}

%\vskip3pt

%\bio{}
%Author biography without author photo.
%Author biography. Author biography. Author biography.
%\endbio

%\bio{figs/cas-pic1}
%Author biography with author photo.
%Author biography. Author biography. Author biography.
%\endbio

%\vskip3pc

%\bio{figs/cas-pic1}
%Author biography with author photo.
%Author biography. Author biography. Author biography.
%\endbio

\end{document}